\documentclass[conference]{IEEEtran}
\IEEEoverridecommandlockouts
\usepackage{cite}
\usepackage{amsmath,amssymb,amsfonts}
\usepackage{algorithmic}
\usepackage{graphicx}
\usepackage{textcomp}
\usepackage{xcolor}
\def\BibTeX{{\rm B\kern-.05em{\sc i\kern-.025em b}\kern-.08em
    T\kern-.1667em\lower.7ex\hbox{E}\kern-.125emX}}

\usepackage{graphics}
\usepackage{url}
\usepackage{multirow}
\usepackage{siunitx}
\usepackage{bm}

\usepackage{adjustbox}
\usepackage{diagbox}

\usepackage[switch]{lineno}

\usepackage{color}

\begin{document}

\title{Inpainting-Driven Mask Optimization \\for Object Removal
\vspace*{-1mm}
}

\author{\IEEEauthorblockN{Anonymous Authors}}

\author{\IEEEauthorblockN{Kodai Shimosato}
\IEEEauthorblockA{\textit{Toyota Technological Institute}, Nagoya, Japan \\
kodai.shimosato.ttij@gmail.com}
\and
\IEEEauthorblockN{Norimichi Ukita}
\IEEEauthorblockA{\textit{Toyota Technological Institute}, Nagoya, Japan \\
ukita@toyota-ti.ac.jp}
}

\maketitle

\vspace*{-8mm}

\begin{abstract}
This paper proposes a mask optimization method for improving the quality of object removal using image inpainting.
While many inpainting methods are trained with a set of random masks, a target for inpainting may be an object, such as a person, in many realistic scenarios.
This domain gap between masks in training and inference images increases the difficulty of the inpainting task.
In our method, this domain gap is resolved by training the inpainting network with object masks extracted by segmentation, and such object masks are also used in the inference step.
Furthermore, to optimize the object masks for inpainting, the segmentation network is connected to the inpainting network and end-to-end trained to improve the inpainting performance.
The effect of this end-to-end training is further enhanced by our mask expansion loss for achieving the trade-off between large and small masks.
Experimental results demonstrate the effectiveness of our method
for better object removal using image inpainting.
\end{abstract}


\section{Introduction}
\label{section:introduction}

Image inpainting fills missing regions, which are provided as a binary mask image, in an image.
In inference, mask regions are given manually (e.g., as lines~\cite{YangWTCY07} and points~\cite{FaridMG16,FaridLG18}) in general.
However, several specific objects tend to be removed and filled in many realistic tasks.
For example, object removal~\cite{DBLP:conf/aaai/ZhangLWGF0YCXY20,DBLP:conf/nips/ShettyFS18} is one of the most popular tasks.

In the training step of image inpainting, on the other hand, most models are trained with random masks~\cite{NazeriNJQE19,DBLP:conf/eccv/ZengLYZSL20,DBLP:conf/cvpr/LiWZDT20,DBLP:conf/cvpr/YiTAJX20,DBLP:conf/iccv/JoP19,DBLP:conf/iccv/ChangLLH19,DBLP:conf/iccv/LiHZDT19,DBLP:conf/iccv/RenYZLLL19,DBLP:conf/iccv/LiuJX019,DBLP:conf/cvpr/WangTSJ19,DBLP:conf/cvpr/XiongYLYLBL19}.
However, the domain gap between masks between the training and inpainting steps drops the inpainting performance. 
To fill this domain gap, this paper focuses on inpainting scenarios in which specific object types, such as people and vehicles, are regarded as inpainting targets.

As with the above domain gap, our method fills the domain gap between object segments extracted by segmentation and object masks optimized for inpainting.
This domain gap is observed because of the difference between the definitions of better segments in segmentation and inpainting.
While a correct object boundary should be extracted in segmentation, the mask should lead to natural image synthesis in inpainting.

\begin{figure}[t]
\vspace*{-3mm}
\begin{center}
\includegraphics[width=1.0\columnwidth]{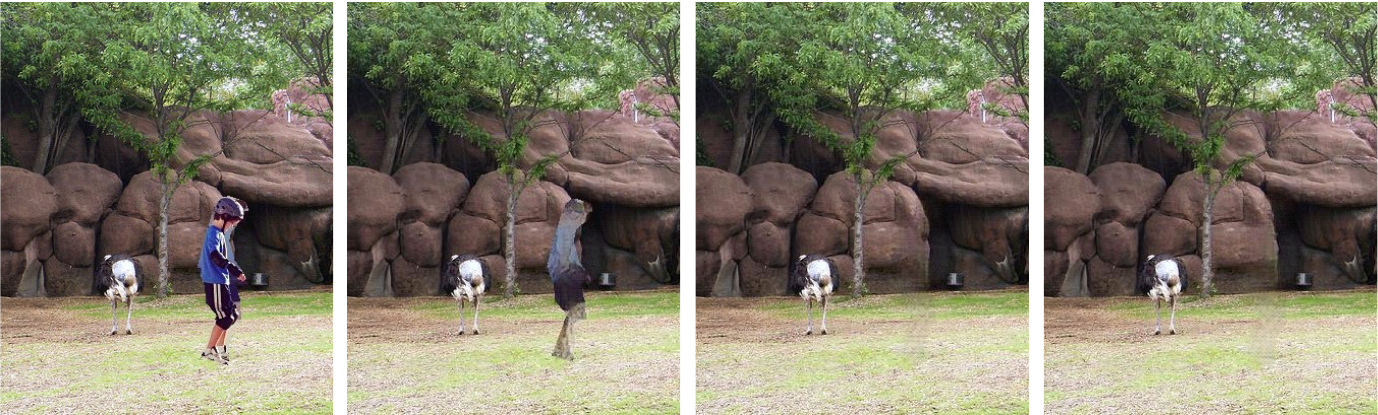}
\end{center}
\begin{minipage}{0.24\columnwidth}
\center{ (a) Input \\image}
\end{minipage}
\begin{minipage}{0.24\columnwidth}
\center{(b) Result \\($-2$ pixels)}
\end{minipage}
\begin{minipage}{0.24\columnwidth}
\center{(c) Result \\($\pm0$ pixels)}
\end{minipage}
\begin{minipage}{0.24\columnwidth}
\center{(d) Result \\($+2$ pixels)}
\end{minipage}
\caption{Inpainting variability occurred by minor mask differences. 
The mask used for reconstructing (c) is the ground-truth segment of a standing person observed in (a).
The mask is eroded and dilated to get (b) and (d), respectively.
}
\label{fig:top}
\end{figure}

To further improve the above object removal, this paper focuses on the effect of the mask shape on image inpainting.
We found that minor shape differences in the mask shape cause a non-negligible variability in inpainted images, as demonstrated in Fig.~\ref{fig:top}.
In this example, a person in (a) is removed and inpainted.
As shown in (b), if the mask is smaller than the object, the object foreground pixels hanging out of the mask boundary remain in the inpainted image and significantly degrade its quality.
On the other hand, even if the mask hangs out of the object boundary as shown in (d), the inpainting quality is not almost changed from the ideal case (c).

Based on the above discussions, our contributions are summarized as follows:
\begin{itemize}
\item For object removal, the domain gap between masks in training and inference images is resolved with training images synthesized by copy-pasting
object regions onto background images.
\item To fill the domain gap between better object segments for the segmentation and inpainting networks, these networks are trained in an end-to-end manner.
\item For further optimizing the mask, our mask expansion loss minimally expands it to include the object region.
\end{itemize}


\section{Related Work}
\label{section:related}

\paragraph{Image Segmentation}

Image segmentation methods~\cite{DBLP:journals/corr/abs-2001-05566} are categorized into semantic segmentation~\cite{DBLP:journals/pami/ShelhamerLD17}, instance segmentation~\cite{DBLP:journals/pami/HeGDG20}, and panoptic segmentation~\cite{DBLP:conf/cvpr/KirillovHGRD19}.
For extracting several foreground object regions as masks for inpainting, instance segmentation is enough.
However, all segments, including the background pixels in an image, are useful to assist inpainting as auxiliary cues, as demonstrated in~\cite{SongYSWHK18,DBLP:conf/eccv/LiaoXWLS20}.
Therefore, for both of these mask extraction and inpainting assistance processes, all pixels in each image are segmented by panoptic segmentation~\cite{DBLP:conf/cvpr/KirillovHGRD19,WangZAYC21} in our method.

\paragraph{Image Inpainting}

It is difficult to reconstruct all complex textures in a wide mask region.
To resolve this problem, various inpainting methods are proposed~\cite{NazeriNJQE19,DBLP:conf/eccv/ZengLYZSL20,DBLP:conf/cvpr/LiWZDT20,DBLP:conf/cvpr/YiTAJX20,DBLP:conf/iccv/JoP19,DBLP:conf/iccv/ChangLLH19,DBLP:conf/iccv/LiHZDT19,DBLP:conf/iccv/RenYZLLL19,DBLP:conf/iccv/LiuJX019,DBLP:conf/cvpr/WangTSJ19,DBLP:conf/cvpr/XiongYLYLBL19}.
For example, such degradation can be relieved with generative adversarial networks (GANs), as demonstrated in~\cite{PathakKDDE16}.

Another way to alleviate the inpainting difficulty is to first reconstruct auxiliary images, which are easier to reconstruct than complex textures in an RGB image. The auxiliary images support RGB image inpainting. 
For example, segmentation~\cite{SongYSWHK18,DBLP:conf/eccv/LiaoXWLS20}, edge~\cite{NazeriNJQE19,DBLP:conf/aaai/YangQS20}, and depth~\cite{yamashita2021} cues are useful.

\paragraph{Masking for Image Inpainting}

In all inpainting methods mentioned above, masks are given by a user.
In~\cite{DBLP:conf/nips/ShettyFS18}, on the other hand, generic foreground objects such as persons are automatically extracted as masks for inpainting.
This inpainting network, including the mask generator, is trained so that foreground object extraction is trained without supervising any object region.
While this method~\cite{DBLP:conf/nips/ShettyFS18} focuses on how to reduce the annotation cost by roughly extracting object regions, we improve inpainting accuracy with more detailed mask optimization.
Object masks are also used in~\cite{DBLP:conf/eccv/ZhengLLCSBZXAL22}.
Furthermore, these masks are dilated to avoid leaking pixels around their boundaries in~\cite{DBLP:conf/eccv/ZhengLLCSBZXAL22}, as with our method.
Unlike constant dilation~\cite{DBLP:conf/eccv/ZhengLLCSBZXAL22}, however, our method further optimizes the masks by end-to-end training and our mask expansion loss.


\section{Inpainting-Driven Mask Optimization by Joint Segmentation-Inpainting Learning}
\label{section:method}

Our joint segmentation-inpainting network is shown in Fig.~\ref{fig:proposed_method}.
Following the success of prior work of joint learning with low- and high-level vision tasks~\cite{DBLP:journals/access/AkitaU23,DBLP:conf/iconip/HarisSU21,DBLP:conf/mva/KondoU21,DBLP:journals/corr/abs-2302-12491}, the motivation of our joint learning is that the segmentation network is trained to improve the inpainting quality because the inpainting result strongly depends on the mask,
as validated in Fig.~\ref{fig:top}.

\subsection{Paired Image Dataset Generation for Supervised Object-specific Inpainting}
\label{subsection:method_dataset}

For object-specific inpainting, pairs of the same images with and without foreground objects are required for supervised inpainting learning.
In general, 
each training image pair is generated so that an original image is randomly masked.
These original and masked images are used as the ground-truth image for supervision and the input image fed into the inpainting model, respectively.
In our method, on the other hand, object regions are superimposed on a background image.

Given images with segmentation annotations, such as the COCO dataset, paired images with and without foreground objects are generated as follows.
In all the original images, object regions are extracted based on their given annotations.
From these extracted object regions, one object region is randomly selected.
This selected region is superimposed by~\cite{DwibediMH17} in a random location on an image selected randomly from the original images.
These randomly selected and superimposed images (i.e., images without and with the superimposed object, respectively) are paired.
In addition, the binary mask image is generated so that the region of the superimposed object is masked.
By collecting a set of these image triplets (i.e., images with and without the superimposed object and its mask image),
the training imageset for our method is generated.

The paired image generation mentioned above is also applied to images in the test imageset.
While only the image with the superimposed object is required in the test step, the generated paired images are used for evaluation.

\subsection{Pretraining}
\label{subsection:method_pretrain}

\begin{figure*}[t]
\begin{center}
\includegraphics[width=\textwidth]{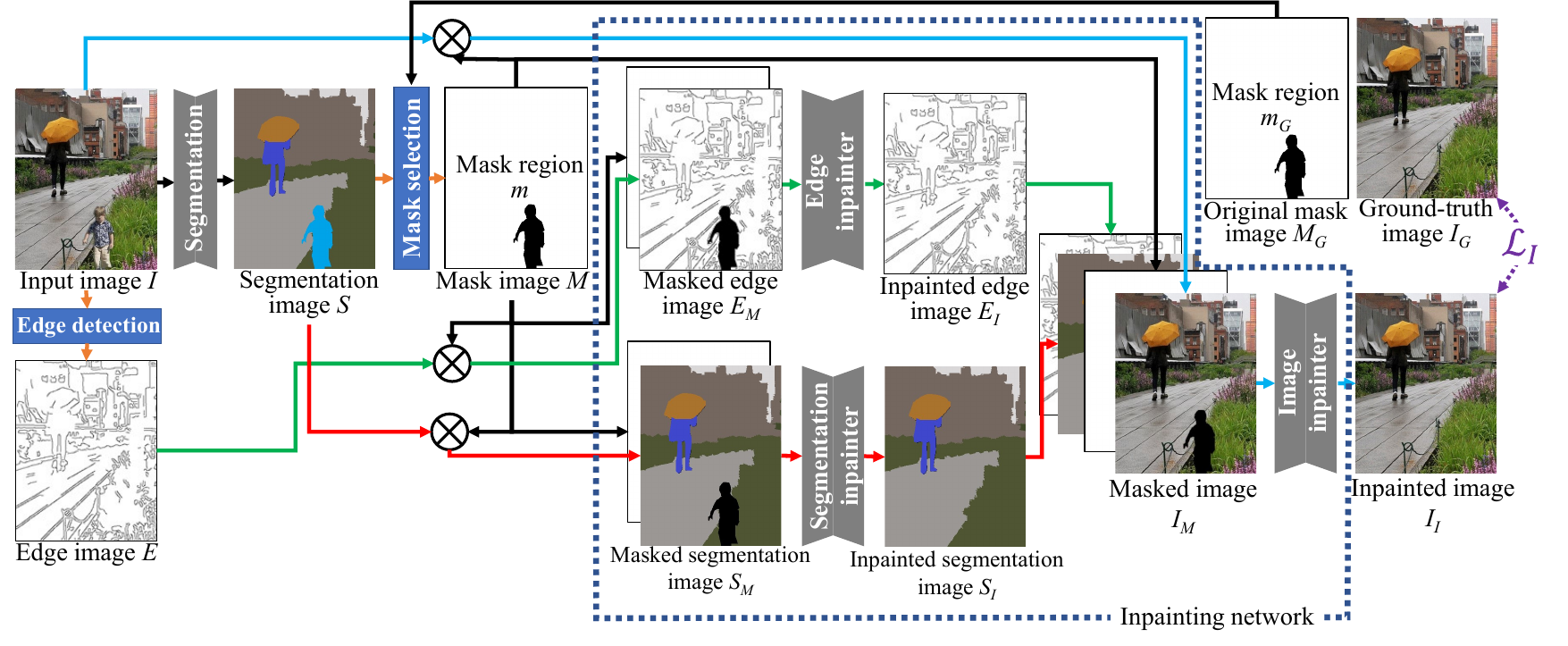}
\caption{Proposed joint segmentation-inpainting network.
Blue, green, red, and black arrows indicate the flows of RGB, edge, segmentation, and mask images, respectively.
Blue rectangles and gray hourglass-shaped boxes indicate processes and learnable sub-networks, respectively.
The inpainting network consists of three inpainting networks, namely the edge, segmentation, and image inpainters, as enclosed by the dotted line.
In training and inference steps, ``$I$, $I_{G}$, and $M_{G}$'' and ``$I$'' are given to this joint network, respectively. 
}
\label{fig:proposed_method}
\end{center}
\end{figure*}

\subsubsection{Pretraining of Segmentation Network}

A panoptic segmentation network is pretrained with its loss function denoted by $\mathcal{L}_{PS}$.
While any differentiable network is employed in our method, PanopticFPN~\cite{DBLP:conf/cvpr/KirillovHGRD19} is used with its pretrained model~\cite{wu2019detectron2} in our experiments.

\subsubsection{Pretraining of Inpainting Network}

As with segmentation, inpainting can be implemented with any differentiable network, while the EdgeConnect-based network introduced in what follows is used in this paper.

From the training paired imageset generated in Sec.~\ref{subsection:method_dataset}, a pair of images with and without a superimposed object (denoted by $I$ and $I_{G}$, respectively) and the original mask image of this object region are given.
This original mask image is denoted by $M_{G}$, and the masked object region in $M_{G}$ is denoted by $m_{G}$.
Given $I$, $I_{G}$ and $M_{G}$, our inpainting network is pretrained as follows.


$I$ is masked by $M_{G}$.
While the main task is to inpaint the masked input image ($I_{M}$), its edge image ($E_{M}$) is
first
inpainted to utilize its inpainted image (denoted by $E_{I}$) as an auxiliary cue for the main task, as done in~\cite{NazeriNJQE19}.
$E_{I}$ is obtained by the Canny detector in the same manner as~\cite{NazeriNJQE19}.
In addition to the edge cue,
the segmentation cue~\cite{SongYSWHK18,DBLP:conf/eccv/LiaoXWLS20} is also used in our method.
The segmentation and its inpainted images are denoted by $S_{M}$ and $S_{I}$, respectively.

The edge inpainter is trained with the weighted sum of the hinge-variant of GAN loss ($\mathcal{L}_{EG}$) and the feature-matching loss~\cite{DBLP:conf/nips/GatysEB15,DBLP:conf/eccv/JohnsonAF16} ($\mathcal{L}_{EF}$):
\begin{eqnarray}
  \mathcal{L}_{E} &=& \lambda_{EG} \mathcal{L}_{EG} + \lambda_{EF} \mathcal{L}_{EF},
  \label{eq:loss_ec_e}\\
  \mathcal{L}_{EG} &=& - D_{E} (E_{I}, I_{G}),
  \label{eq:loss_eg}\\
  \mathcal{L}_{EF} &=& \sum_{i} \frac{1}{N^{i}} \parallel D^{i}_{E} (E_{G})
                      - D^{i}_{E} (E_{I}) \parallel_{1},
  \label{eq:loss_lef}
\end{eqnarray}
where $\lambda_{EG}$ and $\lambda_{EF}$ denote weight constants.
The discriminator $D_{E}$ evaluates whether or not $E_{I}$ is realistic as the edge image of $I_{G}$.
$\mathcal{L}_{EF}$ compares the activation maps in the intermediate layers of $D_{E}$. 
$N^{i}$ and $D^{i}_{E}$ denote the number of elements and the activation in the $i$-th activation layer of $D_{E}$, respectively.

As with the edge inpainter, the segmentation inpainter is trained with the following loss:
\begin{eqnarray}
\mathcal{L}_{S}&=&\lambda_\mathrm{SG}\mathcal{L}_\mathrm{SG} + \lambda_\mathrm{SC}\mathcal{L}_\mathrm{SC},
\label{eq:loss_seg}
\end{eqnarray}
where $\lambda_{SG}$ and $\lambda_{SC}$ denote weight constants.
$\mathcal{L}_\mathrm{SG}$ and $\mathcal{L}_\mathrm{SC}$ denote the hinge-variant of GAN loss, which is equal to Eq. (\ref{eq:loss_eg}), and the cross entropy loss between the pixelwise segmentation labels of $S_{I}$ and its ground-truth.

The inpainted edge ($E_{I}$), inpainted segmentation ($S_{I}$), mask ($M_{G}$), and masked input ($I_{M}$) images are fed into the image inpainter.
This image inpainter is trained with the weighted sum of the hinge-variant of GAN loss ($\mathcal{L}_{IG}$), the feature-matching loss ($\mathcal{L}_{IF}$), the style loss ($\mathcal{L}_{IS}$), and the reconstruction loss ($\mathcal{L}_{IR}$):
\begin{eqnarray}
  \mathcal{L}_{I} &=& \lambda_{IG} \mathcal{L}_{IG} + \lambda_{IF}
                      \mathcal{L}_{IF} + \lambda_{IS} \mathcal{L}_{IS} + \lambda_{IR} \mathcal{L}_{IR},
  \label{eq:loss_ec_i}
\end{eqnarray}
where $\lambda_{IG}$, $\lambda_{IF}$, $\lambda_{IS}$, and $\lambda_{IR}$ are weight constants.
The activations used in $\mathcal{L}_{IF}$ is also employed for $\mathcal{L}_{IS}$ as follows:
\begin{eqnarray}
  \mathcal{L}_{IS} &=& \sum_{i} \parallel G^{i}_{\phi} (I)
                      - G^{i}_{\phi} (I_{I}) \parallel_{1},
  \label{eq:loss_is}
\end{eqnarray}
where $G^{i}_{\phi}$ denotes a Gram matrix constructed from activations $\phi^{i}$~\cite{DBLP:conf/cvpr/GatysEB16}.  
$\mathcal{L}_{IR}$
is the difference between $I$ and $I_{I}$:
\begin{eqnarray}
  \mathcal{L}_{IR} &=& \frac{1}{N_{M}} \parallel I - I_{I} \parallel_{1}
  \label{eq:loss_ir}
\end{eqnarray}
$N_{M}$ denotes the number of the masked pixels $m_{G}$ in $M_{G}$.

In our method, the edge inpainting, segmentation inpainting, and image inpainting networks are independently pretrained with $\mathcal{L}_{E}$ (\ref{eq:loss_ec_e}), $\mathcal{L}_{S}$ (\ref{eq:loss_seg}), and $\mathcal{L}_{I}$ (\ref{eq:loss_ec_i}), respectively.
The architecture of each of these three networks is equal to the base inpainting network presented in~\cite{NazeriNJQE19}.

\subsection{End-to-end Training}
\label{subsection:end_to_end}

The pretrained segmentation and inpainting networks, which are provided as described in Sec.~\ref{subsection:method_pretrain}, are connected as illustrated in Fig.~\ref{fig:proposed_method} for training the connected joint-learning network in an end-to-end manner.
This joint-learning network is trained with all losses of the two networks (i.e., $\mathcal{L}_{PS}$, $\mathcal{L}_{E}$, $\mathcal{L}_{S}$, and $\mathcal{L}_{I}$) as follows.

The input image $I$ is fed into the panoptic segmentation network~\cite{DBLP:conf/cvpr/KirillovHGRD19}, and this network is trained with $\mathcal{L}_{PS}$.
While there are many segments in the segmentation image $S$, only the segment corresponding to $m_{G}$ is selected and fed into the inpainting network.
In this mask selection process of the training step, the segment whose overlap with $m_{G}$ is the largest is selected.
This selected segment (denoted by $m$) is used for making the mask image (denoted by $M$) so that only
$m$ are masked.
For the training of the segmentation network, $M$ is feed-forwarded to the inpainting network, and the inpainting losses (i.e., $\mathcal{L}_{E}$, $\mathcal{L}_{S}$, and $\mathcal{L}_{I}$) are back-propagated through the segmentation network as well as the inpainting network in an end-to-end learning manner.

\subsection{Mask Expansion Loss}
\label{subsection:method_mask}

\subsubsection{Preliminary Experiments}

\begin{table}[t]
\caption{Effect of the mask size on the image inpainting performance. In all tables in this paper, the best score in each row is colored by \textcolor{red}{\textbf{red}}.
For this experiment, only the inpainting network is used without the segmentation network.
}
\label{table:expt_3}
\begin{center}
\begin{footnotesize}
\begin{tabular}{c|c|cccccc}
\hline
\multirow{2}{*}{} & Mask& \multicolumn{6}{c}{Mask size variation (pixels)} \\
 & ratio& -2 & 0 & 2 & 4 & 6 & 8 \\ \hline\hline
{\multirow{3}{*}{\rotatebox[origin=c]{90}{PSNR$\uparrow$}}}
& 0-10 & 26.75 & \textcolor{red}{\textbf{32.31}} & 31.89 & 31.47 & 31.07 & 30.70 \\
 & 10--20 & 17.98 & \textcolor{red}{\textbf{24.77}} & 24.35 & 23.91 & 23.56 & 23.21 \\
 & 20--30 & 16.16 & \textcolor{red}{\textbf{21.45}} & 21.07 & 20.80 & 20.46 & 20.18 \\
 & 30--40 & 14.44 & \textcolor{red}{\textbf{19.21}} & 18.96 & 18.63 & 18.35 & 18.07 \\
\hline
{\multirow{3}{*}{\rotatebox[origin=c]{90}{SSIM$\uparrow$}}}
& 0--10 & 0.943 & \textcolor{red}{\textbf{0.969}} & 0.967 & 0.964 & 0.961 & 0.959 \\
 & 10--20 & 0.807 & \textcolor{red}{\textbf{0.892}} & 0.883 & 0.875 & 0.866 & 0.858 \\
 & 20--30 & 0.704 & \textcolor{red}{\textbf{0.807}} & 0.796 & 0.785 & 0.773 & 0.762 \\
 & 30--40 & 0.599 & \textcolor{red}{\textbf{0.711}} & 0.700 & 0.687 & 0.675 & 0.662 \\
\hline
\end{tabular}
\end{footnotesize}
\end{center}
\end{table}

In addition to Fig.~\ref{fig:top}, this section further verifies the effect of the mask size on the inpainting performance, while a similar effect is found by using only a specific amount of dilation given to an inpainting mask in~\cite{DBLP:conf/eccv/ZhengLLCSBZXAL22}.
The detailed conditions of all experiments shown in this paper are described in Sec.~\ref{subsection:exp_details}.
Table~\ref{table:expt_3} shows how the inpainting quality changes depending on the dilation and erosion of the mask.
The ground-truth mask of each test image is dilated and eroded by a fixed distance $d \in \{ -2, 0, 2, 4, 6, 8 \}$ pixels, where the positive and negative distances are the outside and inside of the mask boundary, respectively.
To verify the performance in detail, test images are divided into four groups
depending on the percentage of mask pixels in each test image.

The best result is obtained in the original ground-truth mask in all mask ratios of all metrics.
However, it is extremely difficult to conform the extracted mask to its ground truth.
Next, compare the scores of the masks rescaled with 2 and -2 pixels.
While the absolute distances of these two rescales are the same, the performance drop with the mask rescaled with -2 pixels is much larger.
It can also be seen that the score slowly degrades as the mask size increases.
Based on this observation, it is concluded that a trade-off between large and small masks is important.
For the trade-off for better inpainting, our proposed method optimizes $M$ so that $m$ is expanded minimally enough to cover $m_{G}$.


$M$ is determined by the segmentation network in our joint-learning network shown in Fig.~\ref{fig:proposed_method}.
In Secs.~\ref{subsection:method_pretrain} and \ref{subsection:end_to_end}, this network is trained with two criteria, namely (i) the segmentation image $S$ is the same as its ground-truth and (ii) the inpainted image $I_{I}$ is the same as its ground-truth $I_{G}$.
The first criterion achieved with $\mathcal{L}_{PS}$ does not encourage $M$ to be larger than $M_{G}$.
In the second criterion achieved with $\mathcal{L}_{I}$, $m$ may become larger than $m_{G}$ to improve the inpainting quality.
However, no explicit force for expanding $m$ is given in the second criterion.

For explicitly expanding $m$ so that $m$ covers $m_{G}$ in the segmentation network, we propose to train the segmentation network with one more loss.
As with $\mathcal{L}_{PS}$ of PanopticFPN~\cite{DBLP:conf/cvpr/KirillovHGRD19}, many segmentation networks~\cite{DBLP:conf/cvpr/KirillovWHG20,kwon2022segmentation} use region-based loss functions such as the cross entropy loss.
Since the region-based loss evaluates the pixelwise class labels of all pixels, it is not good at locally adjusting the boundary of a mask~\cite{DBLP:journals/mia/MaCNHLLYM21}.
For directly adjusting the local boundary of $m$, our method uses a boundary-based loss.

\subsubsection{Boundary Loss}

The Boundary loss $\mathcal{L}_{B}$~\cite{KervadecBDGDA19} is the distance-weighted 2D area between the ground-truth region (i.e., $m_{G}$) and its estimated one, which becomes zero in the ideal estimation:
\begin{equation}
\mathcal{L}_{B} = \int_{\Omega} \phi_{m_{G}} (p) \hat{m}_\theta (p) dp, \label{eq:boundary_loss}
\end{equation}
where $p$, $\Omega$, and $\hat{m}_\theta (p)$ denote a point on the boundary of $m_{G}$, a pixel set in the image, and the softmax output of the segmentation network at $p$.
$\phi_{m_{G}} (p) = -D_{m_{G}}(p)$ if $p \in m_{G}$, and otherwise $\phi_{m_{G}} (p) = D_{m_{G}}(p)$, where $D_{m_{G}}(p)$ is the distance map from the boundary of $m_{G}$.

\subsubsection{Mask Expansion Loss}

$\mathcal{L}_{B}$ is designed to conform the boundary of the predicted region to that of its ground truth.
With $\mathcal{L}_{B}$, our mask expansion loss is implemented as follows.
Since $\phi_{m_{G}} (p)$ in Eq.~(\ref{eq:boundary_loss}) is a negative distance value as defined to be $\phi_{m_{G}} (p) = -D_{m_{G}}(p)$ if $p \notin m_{G}$, the boundary of the estimated mask (i.e., $m$) moves outside of $m_{G}$ by decreasing $\phi_{m_{G}} (p)$.
As the result of moving $m$ to the outside of $m_{G}$, $m$ is expanded so that $m$ includes $m_{G}$.

\begin{figure}[t]
\begin{center}
\includegraphics[width=\columnwidth]{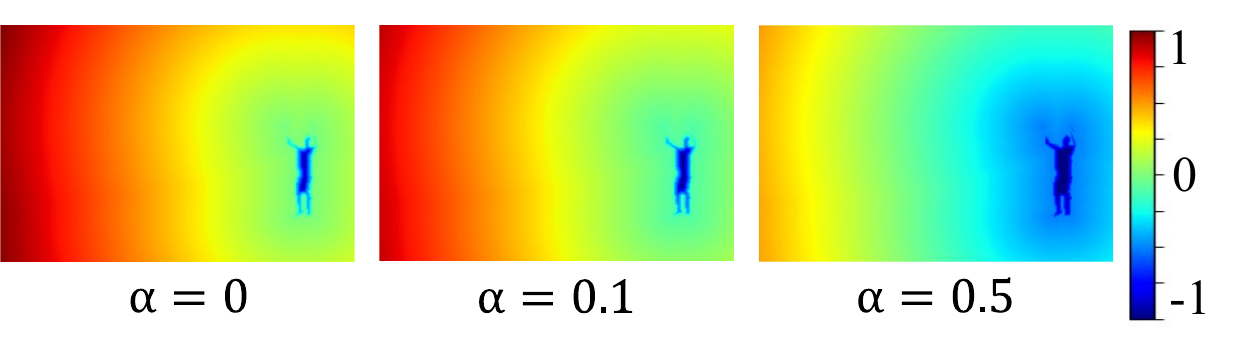}
\caption{Distance maps of different $\alpha$ in Eq.~(\ref{eq:constant_expansion_loss}). The distance values are normalized between -1 and 1.}
\label{fig:bl_offset}
\end{center}
\end{figure}

Based on the above discussion, this paper proposes the following mask expansion loss $\mathcal{L}_{X}$.
\begin{eqnarray}
    \mathcal{L}_{X} (\theta) = \int _{\Omega} \left( \phi_{m_{G}} (p) - \alpha \right) \hat{m}_{\theta} (p) dq,
    \label{eq:constant_expansion_loss}
\end{eqnarray}
where $\alpha$ denotes a constant hyper parameter.
In $\mathcal{L}_{X}$,  $\phi_{m_{G}} (p)$ in Eq.~(\ref{eq:boundary_loss}) is decreased by $\alpha$.
As $\alpha$ gets larger, the boundary line where the distance is zero is dilated (Fig.~\ref{fig:bl_offset}).

With $\mathcal{L}_{X}$, the total loss $\mathcal{L}_{SN}$
is defined as follows:
\begin{eqnarray}
    \mathcal{L}_{SN} = \mathcal{L}_{PS} + \mathcal{L}_{E} + \mathcal{L}_{S} + \mathcal{L}_{I} + \lambda_{X} \mathcal{L}_{X},
    \label{eq:total_segmentation_loss}
\end{eqnarray}
where $\lambda_{X} = 1000$ in our experiments.

\subsection{Inference}
\label{subsubsection:proposed_inference}

In an actual usage scenario, a difference between the
training and inference steps is the mask selection process.
While the original mask image $M_{G}$ is provided to select $m$ in the training step, $M_{G}$ is not available in the inference step.
Instead, $m$ is selected from segments in $S$ by a user.
For our experiments shown in Sec.~\ref{section:experiments}, on the other hand, $M_{G}$ is provided to select $m$ also in the inference step for an automatic and fair comparison.


\begin{table}[t]
\caption{Effect of the training mask type on the image inpainting performance.
For this experiment, only the inpainting network is used without the segmentation network.
}
\label{table:exp_1}
\begin{center}
\begin{tabular}{c|c|cc}
\hline
 & \multirow{2}{*}{Mask ratio}& Random mask & Object mask \\
 & & training & training \\ \hline\hline
 \multirow{3}{*}{PSNR$\uparrow$} & 0--10 & 32.07 & \textcolor{red}{\textbf{32.31}} \\
 & 10--20 & 23.97 & \textcolor{red}{\textbf{24.77}} \\
 & 20--30 & 20.20 & \textcolor{red}{\textbf{21.45}} \\
 & 30--40 & 17.81 & \textcolor{red}{\textbf{19.21}} \\
\hline
 \multirow{3}{*}{SSIM$\uparrow$} & 0--10 & 0.968 & \textcolor{red}{\textbf{0.969}} \\
 & 10--20 & 0.888 & \textcolor{red}{\textbf{0.892}} \\
 & 20--30 & 0.799 & \textcolor{red}{\textbf{0.807}} \\  & 30--40 & 0.702 & \textcolor{red}{\textbf{0.711}} \\
\hline
\end{tabular}
\end{center}
\end{table}
\begin{table}[t]
\caption{Effect of the number of training paired images for fine-tuning from the network trained with random masks, whose results are shown in ``Random mask training'' in Table~\ref{table:exp_1}. 
For this experiment, only the inpainting network is used without the segmentation network.
}
\label{table:exp_2}
\begin{center}
\begin{tabular}{c|c|ccccc}
\hline
& Mask& \multicolumn{4}{c}{\# of images ($\times$ 1k)} \\
 &ratio & 0& 20 & 60 & 100 & 110 \\ \hline\hline
{\multirow{3}{*}{\rotatebox[origin=c]{90}{PSNR$\uparrow$}}}
& 0--10 & 32.07&32.10 & 32.17 & \textcolor{red}{\textbf{32.19}} & 32.12 \\
 & 10--20 & 23.97&24.50 & 24.56 & \textcolor{red}{\textbf{24.59}} & 24.55 \\
 & 20--30 & 20.20&21.17& \textcolor{red}{\textbf{21.22}} & 21.18 & 21.13 \\
 & 30--40 & 17.81&\textcolor{red}{\textbf{18.98}} & 18.77 & 18.94 & 18.79 \\
\hline
{\multirow{3}{*}{\rotatebox[origin=c]{90}{SSIM$\uparrow$}}}
& 0--10 & \textcolor{red}{0.968}&\textcolor{red}{\textbf{0.968}} & \textcolor{red}{\textbf{0.968}} & \textcolor{red}{\textbf{0.968}} & \textcolor{red}{\textbf{0.968}} \\
 & 10--20 & 0.888&0.888 & 0.888 & \textcolor{red}{\textbf{0.889}} & 0.888 \\
 & 20--30 & 0.799&0.800 & \textcolor{red}{\textbf{0.803}} & \textcolor{red}{\textbf{0.803}} & 0.800 \\
 & 30--40 & 0.702&0.703 & 0.705 & \textcolor{red}{\textbf{0.706}} & 0.702 \\
\hline
\end{tabular}
\end{center}
\end{table}

\begin{table}[t]
\caption{Inpainting accuracy change in accompany with change in $\alpha$ of the mask expansion loss (\ref{eq:constant_expansion_loss}). For simplicity, the mean score of all mask ratios is shown in each $\alpha$.
}
\label{table:expt5_2}
\begin{center}
\begin{tabular}{l|cccc}
\hline
 & \multicolumn{4}{c}{$\alpha$} \\
 & 0.01 & 0.03 & 0.05 & 0.07 \\ \hline\hline
PSNR$\uparrow$ & 30.68 & \textcolor{red}{\textbf{30.87}} & 30.51 & 30.47 \\
SSIM$\uparrow$ & 0.951 & \textcolor{red}{\textbf{0.954}} & 0.949 & 0.949 \\
\hline
\end{tabular}
\end{center}
\end{table}

\section{Experimental Results}
\label{section:experiments}

\begin{table*}[t]
\caption{Quantitative comparison with SOTA inpainting methods.
The scores of our method without the Mask Expansion Loss (MEL) is also shown.
The best and second-best scores in each row are colored by \textcolor{red}{\textbf{red}} and \textcolor{blue}{\textbf{blue}}, respectively.}
\label{table:exp_7}
\begin{center}
\begin{tabular}{c|c|ccccccc}
\hline
 & \multirow{1}{*}{Mask ratio} & G\&L & Contextual & Gated & EdgeC & Hi-fill & Bat-fill & Wavefill \\
 \hline\hline
 &  0-10 & 28.48 & 27.43& 26.73& 28.56& \textcolor{blue}{\textbf{30.43}}& 23.29& 25.84 \\
 & 10-20 & 20.38 & 18.59& 18.44& 19.82& 20.60& 19.13& 19.77 \\
 & 20-30 & 18.11 & 16.32& 16.50& 17.72& 17.04& 17.38& 17.97 \\
\multirow{-4}{*}{PSNR$\uparrow$} & 30-40 & 16.21 & 14.67& 14.95& 15.69& 14.24& 15.77& 15.85\\
\hline
 &  0-10 & 0.950 & 0.946& 0.944& 0.950& \textcolor{red}{\textbf{0.958}}& 0.850& 0.942 \\
 & 10-20 & 0.834 & 0.813& 0.816& 0.829& 0.837& 0.751& 0.830 \\
 & 20-30 & 0.747 & 0.716& 0.726& 0.743& 0.718& 0.670& 0.749 \\
\multirow{-4}{*}{SSIM$\uparrow$} & 30-40 & 0.658 & 0.620& 0.637& 0.654& 0.598& 0.593& 0.658 \\
\hline
 &  0-10 & \textcolor{blue}{\textbf{0.069}} & 0.071& 0.074& 0.070& \textcolor{red}{\textbf{0.064}}& 0.407& 0.076 \\
 & 10-20 & 0.167 & 0.168& 0.172& 0.166& \textcolor{red}{\textbf{0.162}}& 0.455& 0.165 \\
 & 20-30 & 0.245 & 0.243& 0.242& 0.237& 0.256& 0.506& 0.236\\
\multirow{-4}{*}{LPIPS$\downarrow$} & 30-40 & 0.331 & 0.323& 0.317& 0.315& 0.350& 0.545& 0.312\\
\hline
 &  0-10 & 0.31& 0.38& 0.51& 0.42& \textcolor{red}{\textbf{0.12}}& 3.11& 0.71 \\
 & 10-20 & 7.48 & 6.33& 6.29& 7.19& 2.19& 5.85& 6.18 \\
 & 20-30 & 11.17 & 8.65& 9.50& 10.54& 6.14& 7.43& 7.81 \\
\multirow{-3}{*}{FID$\downarrow$} & 30-40 & 16.25 & 15.78& 14.30& 14.47& 16.04& 11.37& 11.74 \\
\hline
\hline
 & \multirow{1}{*}{Mask ratio} & RFR& Boundary& ZITS& MAT& LDM & Ours w/o MEL
 & Ours \\
 \hline\hline
 &  0-10 & 29.09& 27.60& 24.71& 23.67& 26.22& 29.60& \textcolor{red}{\textbf{31.07}} \\
 & 10-20 & 19.95& 19.66& 19.01& 15.54& 18.54& \textcolor{blue}{\textbf{24.23}}& \textcolor{red}{\textbf{24.71}} \\
 & 20-30 &17.66& 17.71 & 17.87& 13.18& 16.47& \textcolor{blue}{\textbf{24.66}}& \textcolor{red}{\textbf{25.62}} \\
\multirow{-4}{*}{PSNR$\uparrow$} & 30-40 & 15.28& 15.99& 16.63& 11.50& 14.93& \textcolor{blue}{\textbf{26.13}}& \textcolor{red}{\textbf{26.50}} \\
\hline
 &  0-10 &0.952& 0.947& 0.916& 0.916& 0.930& 0.945& \textcolor{blue}{\textbf{0.955}} \\
 & 10-20 &0.831& 0.828& 0.805& 0.771& 0.806& \textcolor{blue}{\textbf{0.865}}& \textcolor{red}{\textbf{0.874}} \\
 & 20-30 &0.742& 0.744& 0.733& 0.664& 0.713& \textcolor{blue}{\textbf{0.839}}& \textcolor{red}{\textbf{0.848}} \\
\multirow{-4}{*}{SSIM$\uparrow$} & 30-40 &0.645& 0.657& 0.665& 0.564& 0.628& \textcolor{blue}{\textbf{0.844}}& \textcolor{red}{\textbf{0.850}} \\
\hline
 &  0-10 &  \textcolor{blue}{\textbf{0.069}}& 0.073& 0.151& 0.131& 0.122& 0.099& 0.094 \\
 & 10-20 &  \textcolor{blue}{\textbf{0.163}}& 0.170& 0.225& 0.234& 0.207& 0.170& 0.165 \\
 & 20-30 & 0.233& 0.244& 0.279& 0.315& 0.273& \textcolor{blue}{\textbf{0.191}}& \textcolor{red}{\textbf{0.187}}\\
\multirow{-4}{*}{LPIPS$\downarrow$} & 30-40 & 0.310& 0.319& 0.329& 0.393& 0.332& \textcolor{red}{\textbf{0.183}}& \textcolor{blue}{\textbf{0.193}}\\
\hline
 &  0-10 & 0.24& 0.62& 0.64& 1.24& 0.24& \textcolor{blue}{\textbf{0.14}}& 0.18 \\
 & 10-20 & 4.66& 7.70& 5.05& 12.20& 4.82& \textcolor{red}{\textbf{1.75}}& \textcolor{blue}{\textbf{2.02}} \\
 & 20-30 & 7.01& 10.38& 5.37& 18.92& 5.43& \textcolor{red}{\textbf{2.56}}& \textcolor{blue}{\textbf{3.02}} \\
\multirow{-3}{*}{FID$\downarrow$} & 30-40 & 14.48& 14.24& 6.77& 2.522& 10.26& \textcolor{red}{\textbf{1.85}}& \textcolor{blue}{\textbf{2.45}} \\
\hline
\end{tabular}
\end{center}
\end{table*}

\begin{figure*}
\begin{center}
\begin{minipage}{0.16\textwidth}
\begin{center}
 \includegraphics[width=\textwidth]{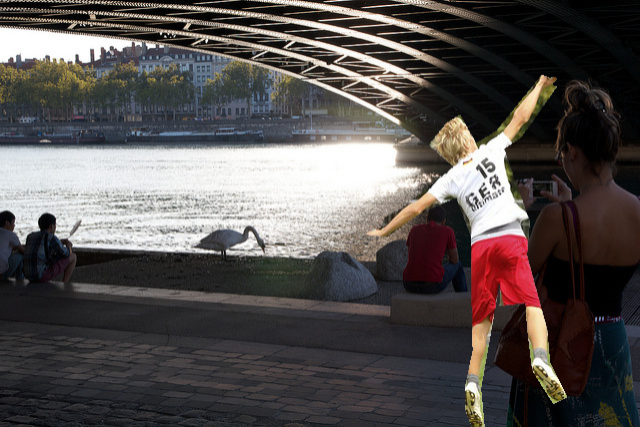}
 (a) Input\\~
\end{center}
\end{minipage}
\begin{minipage}{0.16\textwidth}
\begin{center}
 \includegraphics[width=\textwidth]{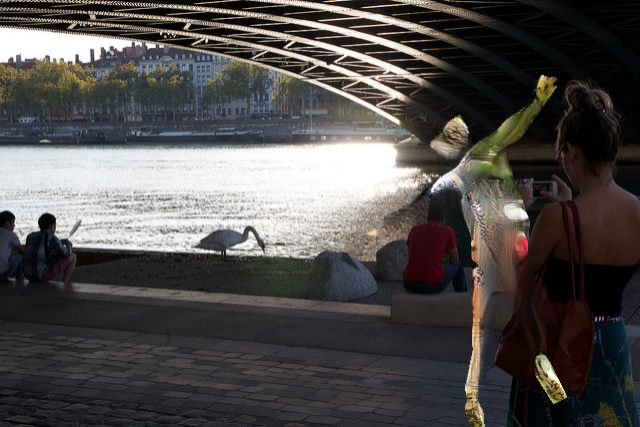}
 (b) G\&L~\cite{DBLP:journals/tog/IizukaS017}
\end{center}
\end{minipage}
\begin{minipage}{0.16\textwidth}
\begin{center}
 \includegraphics[width=\textwidth]{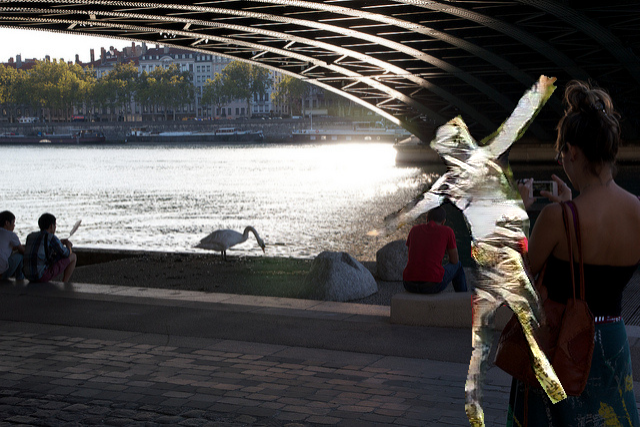}
 (c) Contextual~\cite{DBLP:conf/cvpr/Yu0YSLH18}
\end{center}
\end{minipage}
\begin{minipage}{0.16\textwidth}
\begin{center}
 \includegraphics[width=\textwidth]{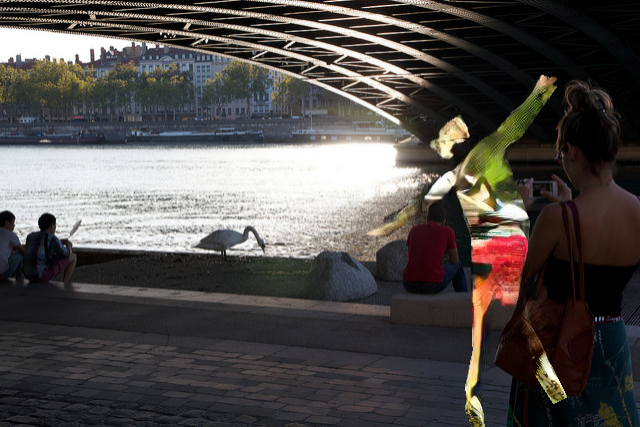}
 (d) Gated~\cite{DBLP:conf/iccv/YuLYSLH19}
\end{center}
\end{minipage}
\begin{minipage}{0.16\textwidth}
\begin{center}
 \includegraphics[width=\textwidth]{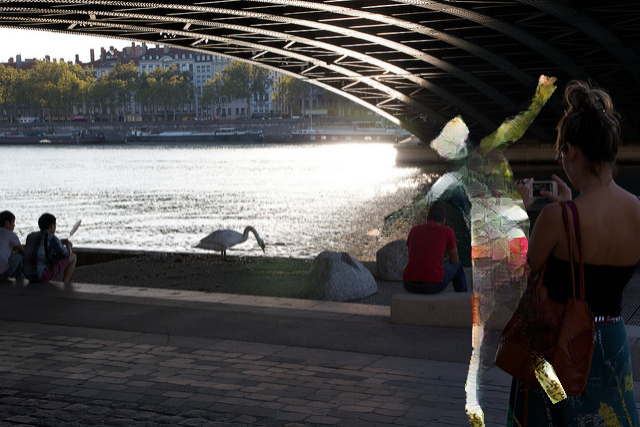}
 (e) EdgeC~\cite{NazeriNJQE19}
\end{center}
\end{minipage}
\begin{minipage}{0.16\textwidth}
\begin{center}
 \includegraphics[width=\textwidth]{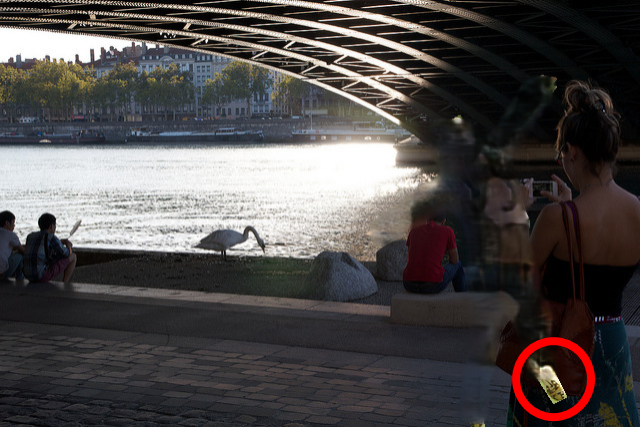}
 (f) Hi-fill~\cite{DBLP:conf/cvpr/YiTAJX20}
\end{center}
\end{minipage}\\
\begin{minipage}{0.16\textwidth}
\begin{center}
 \includegraphics[width=\textwidth]{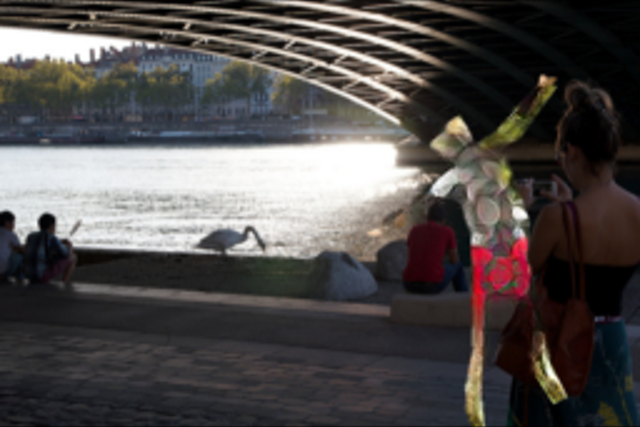}
 (g) Bat-fill~\cite{DBLP:conf/mm/YuZWPCLMXM21}
\end{center}
\end{minipage}
\begin{minipage}{0.16\textwidth}
\begin{center}
 \includegraphics[width=\textwidth]{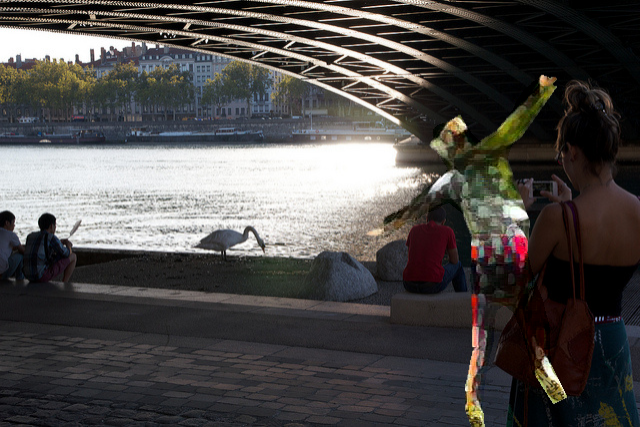}
 (h) Wavefill~\cite{DBLP:conf/iccv/YuZLPMXM21}
\end{center}
\end{minipage}
\begin{minipage}{0.16\textwidth}
\begin{center}
 \includegraphics[width=\textwidth]{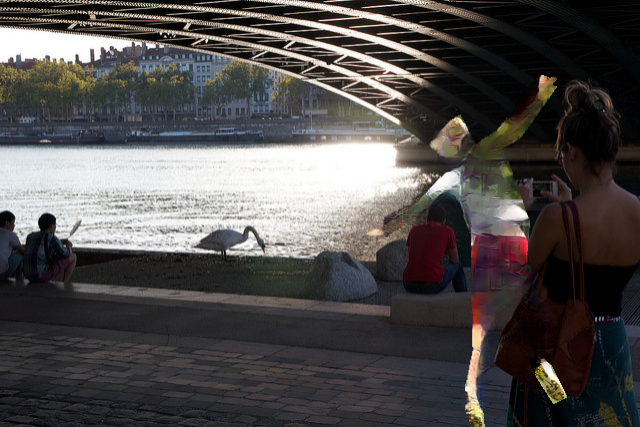}
 (i) Boundary~\cite{yamashita2021}
\end{center}
\end{minipage}
\begin{minipage}{0.16\textwidth}
\begin{center}
 \includegraphics[width=\textwidth]{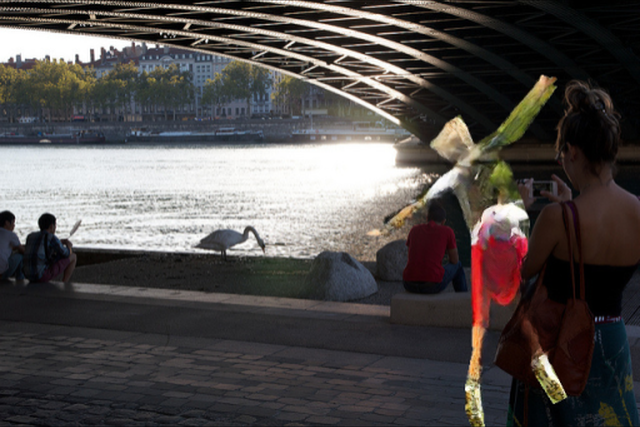}
 (j) ZITS~\cite{DBLP:conf/cvpr/DongCF22}
\end{center}
\end{minipage}
\begin{minipage}{0.16\textwidth}
\begin{center}
 \includegraphics[width=\textwidth]{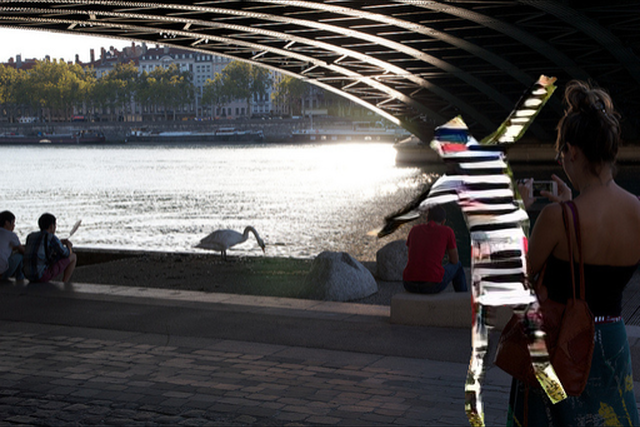}
 (k) MAT~\cite{DBLP:conf/cvpr/LiLZQWJ22}
\end{center}
\end{minipage}
\begin{minipage}{0.16\textwidth}
\begin{center}
 \includegraphics[width=\textwidth]{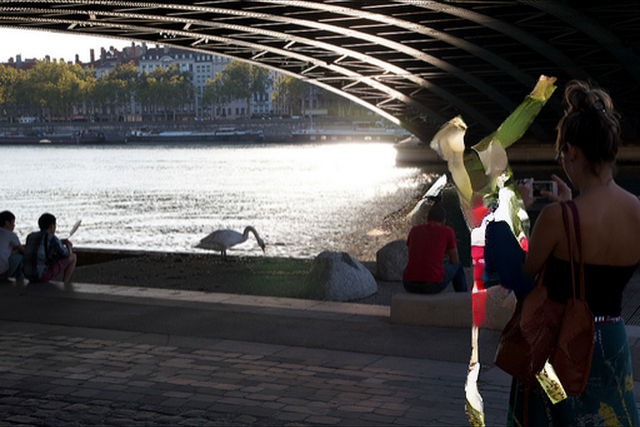}
 (l) LDM~\cite{DBLP:conf/cvpr/RombachBLEO22}
\end{center}
\end{minipage}\\
\begin{minipage}{0.16\textwidth}
\begin{center}
 \includegraphics[width=\textwidth]{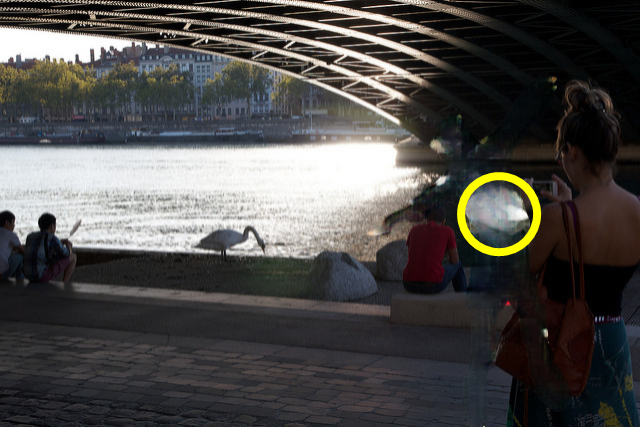}
 (m) Ours w/o MEL
\end{center}
\end{minipage}
\begin{minipage}{0.16\textwidth}
\begin{center}
 \includegraphics[width=\textwidth]{./fig/001268/ours.png}
 (n) Ours
\end{center}
\end{minipage}
\begin{minipage}{0.16\textwidth}
\begin{center}
 \includegraphics[width=\textwidth]{./fig/001268/GT.png}
 (o) GT
\end{center}
\end{minipage}
\begin{minipage}{0.16\textwidth}
\begin{center}
 \includegraphics[width=\textwidth]{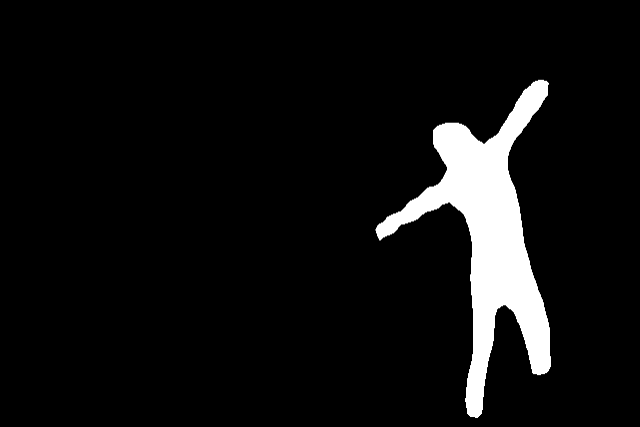}
 (p) Mask
\end{center}
\end{minipage}
\begin{minipage}{0.16\textwidth}
\begin{center}
 \includegraphics[width=\textwidth]{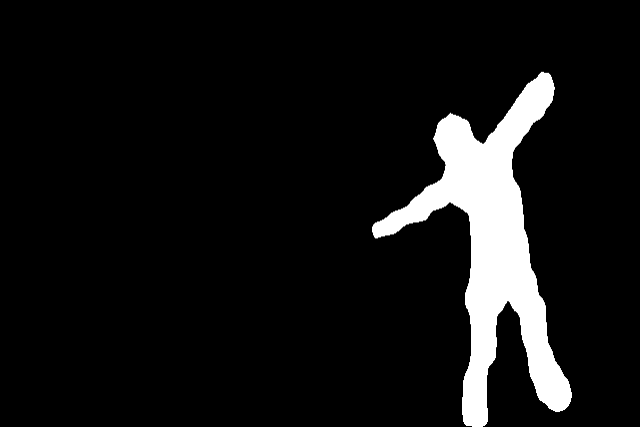}
 (q) Mask by (m)
\end{center}
\end{minipage}
\begin{minipage}{0.16\textwidth}
\begin{center}
 \includegraphics[width=\textwidth]{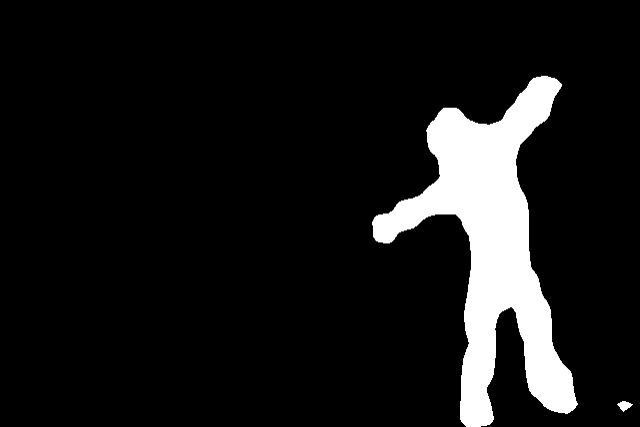}
 (r) Mask by (n)
\end{center}
\end{minipage}\\
\begin{minipage}{0.16\textwidth}
\begin{center}
 \includegraphics[width=\textwidth]{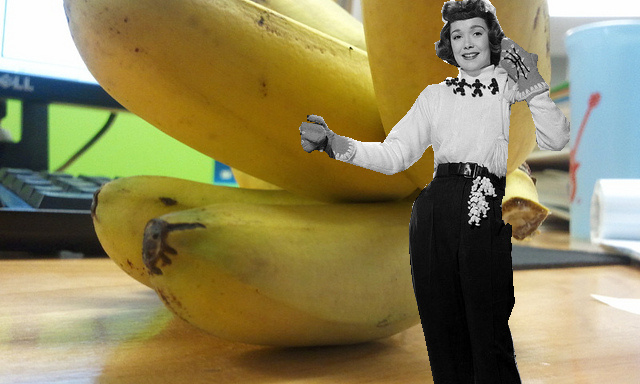}
 (a) Input\\~
\end{center}
\end{minipage}
\begin{minipage}{0.16\textwidth}
\begin{center}
 \includegraphics[width=\textwidth]{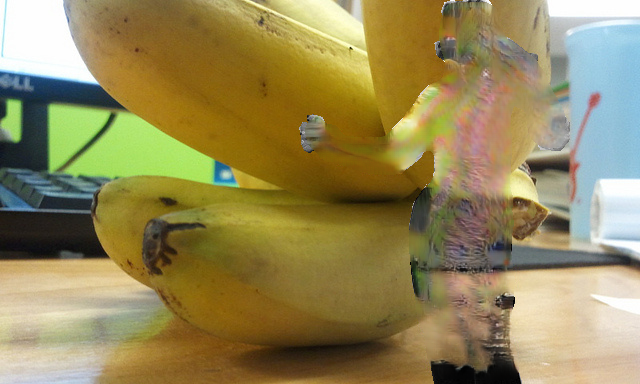}
 (b) G\&L~\cite{DBLP:journals/tog/IizukaS017}
\end{center}
\end{minipage}
\begin{minipage}{0.16\textwidth}
\begin{center}
 \includegraphics[width=\textwidth]{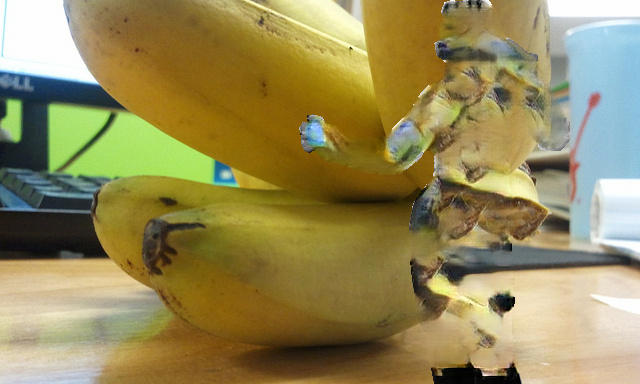}
 (c) Contextual~\cite{DBLP:conf/cvpr/Yu0YSLH18}
\end{center}
\end{minipage}
\begin{minipage}{0.16\textwidth}
\begin{center}
 \includegraphics[width=\textwidth]{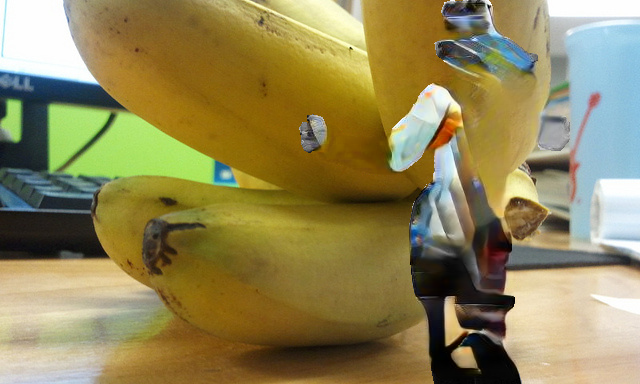}
 (d) Gated~\cite{DBLP:conf/iccv/YuLYSLH19}
\end{center}
\end{minipage}
\begin{minipage}{0.16\textwidth}
\begin{center}
 \includegraphics[width=\textwidth]{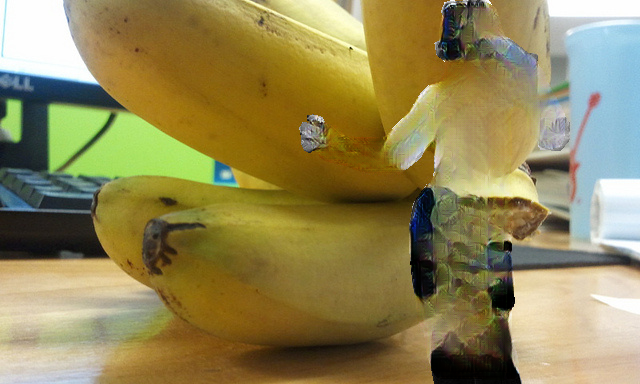}
 (e) EdgeC~\cite{NazeriNJQE19}
\end{center}
\end{minipage}
\begin{minipage}{0.16\textwidth}
\begin{center}
 \includegraphics[width=\textwidth]{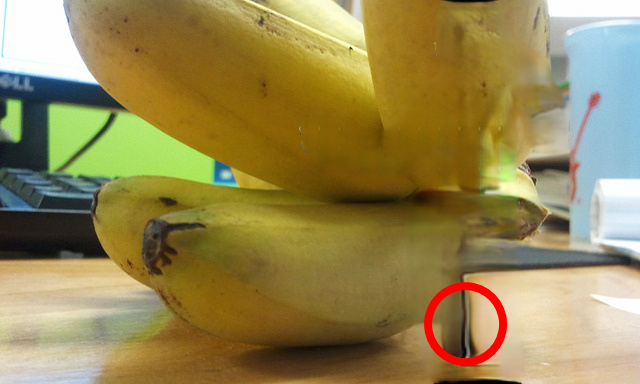}
 (f) Hi-fill~\cite{DBLP:conf/cvpr/YiTAJX20}
\end{center}
\end{minipage}\\
\begin{minipage}{0.16\textwidth}
\begin{center}
 \includegraphics[width=\textwidth]{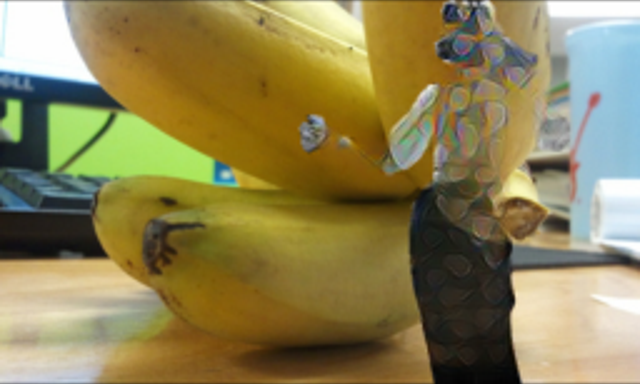}
 (g) Bat-fill~\cite{DBLP:conf/mm/YuZWPCLMXM21}
\end{center}
\end{minipage}
\begin{minipage}{0.16\textwidth}
\begin{center}
 \includegraphics[width=\textwidth]{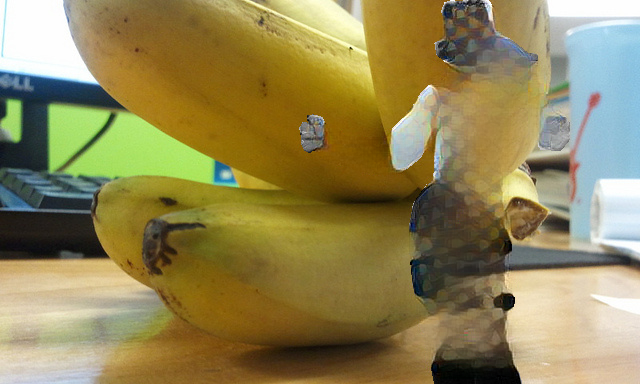}
 (h) Wavefill~\cite{DBLP:conf/iccv/YuZLPMXM21}
\end{center}
\end{minipage}
\begin{minipage}{0.16\textwidth}
\begin{center}
 \includegraphics[width=\textwidth]{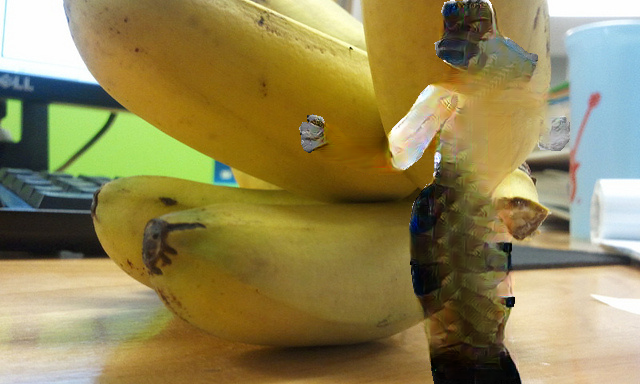}
 (i) Boundary~\cite{yamashita2021}
\end{center}
\end{minipage}
\begin{minipage}{0.16\textwidth}
\begin{center}
 \includegraphics[width=\textwidth]{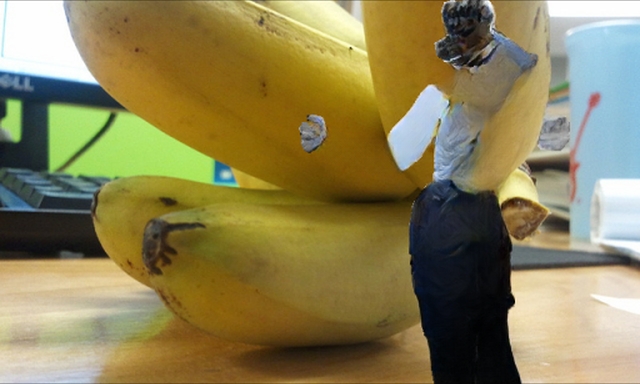}
 (j) ZITS~\cite{DBLP:conf/cvpr/DongCF22}
\end{center}
\end{minipage}
\begin{minipage}{0.16\textwidth}
\begin{center}
 \includegraphics[width=\textwidth]{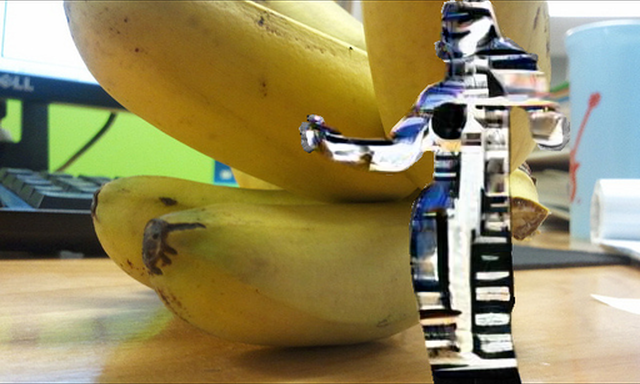}
 (k) MAT~\cite{DBLP:conf/cvpr/LiLZQWJ22}
\end{center}
\end{minipage}
\begin{minipage}{0.16\textwidth}
\begin{center}
 \includegraphics[width=\textwidth]{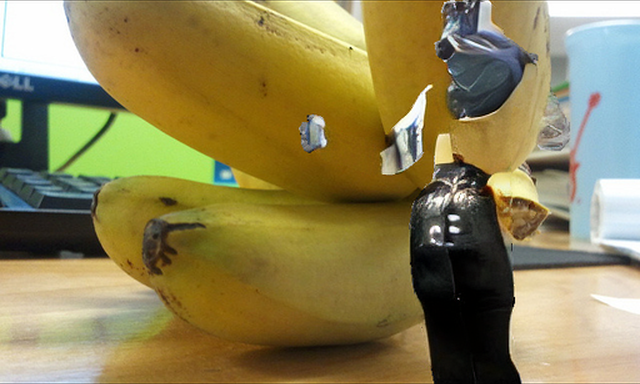}
 (l) LDM~\cite{DBLP:conf/cvpr/RombachBLEO22}
\end{center}
\end{minipage}\\
\begin{minipage}{0.16\textwidth}
\begin{center}
 \includegraphics[width=\textwidth]{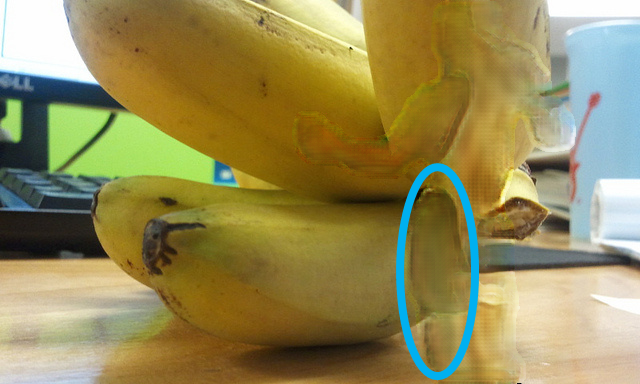}
 (m) Ours w/o MEL
\end{center}
\end{minipage}
\begin{minipage}{0.16\textwidth}
\begin{center}
 \includegraphics[width=\textwidth]{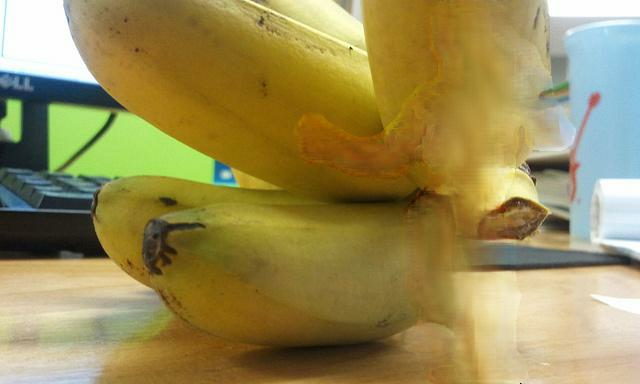}
 (n) Ours
\end{center}
\end{minipage}
\begin{minipage}{0.16\textwidth}
\begin{center}
 \includegraphics[width=\textwidth]{./fig/003661/GT.png}
 (o) GT
\end{center}
\end{minipage}
\begin{minipage}{0.16\textwidth}
\begin{center}
 \includegraphics[width=\textwidth]{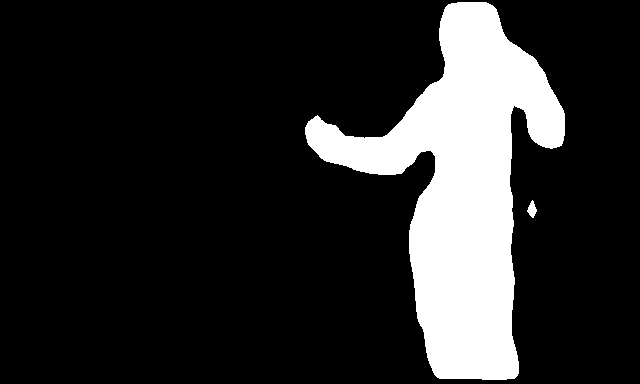}
 (p) Mask
\end{center}
\end{minipage}
\begin{minipage}{0.16\textwidth}
\begin{center}
 \includegraphics[width=\textwidth]{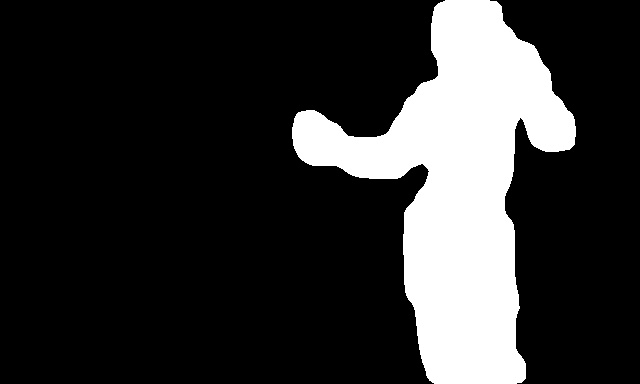}
 (q) Mask by (m)
\end{center}
\end{minipage}
\begin{minipage}{0.16\textwidth}
\begin{center}
 \includegraphics[width=\textwidth]{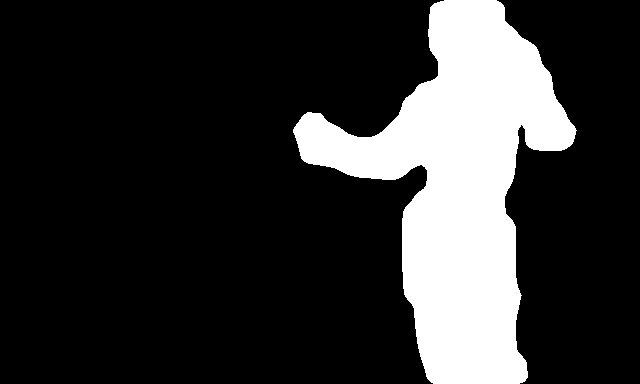}
 (r) Mask by (n)
\end{center}
\end{minipage}\\
\caption{Visual results (success cases). An object region is pasted onto the GT image (l) in order to synthesize the input image (a). This input image is fed into each inpainting network with a mask image. In the previous inpainting methods~\cite{DBLP:journals/tog/IizukaS017,DBLP:conf/cvpr/Yu0YSLH18,DBLP:conf/iccv/YuLYSLH19,NazeriNJQE19,DBLP:conf/cvpr/YiTAJX20,DBLP:conf/mm/YuZWPCLMXM21,DBLP:conf/iccv/YuZLPMXM21,DBLP:conf/cvpr/LiWZDT20,yamashita2021}, 
the mask image (p) generated by PanopticFPN
is used. On the other hand, the mask images (j) and (k) are estimated in our methods.}
 \label{fig:results_expt4_success}
\end{center}
\begin{center}
\begin{minipage}{0.160\textwidth}
\includegraphics[width=\textwidth]{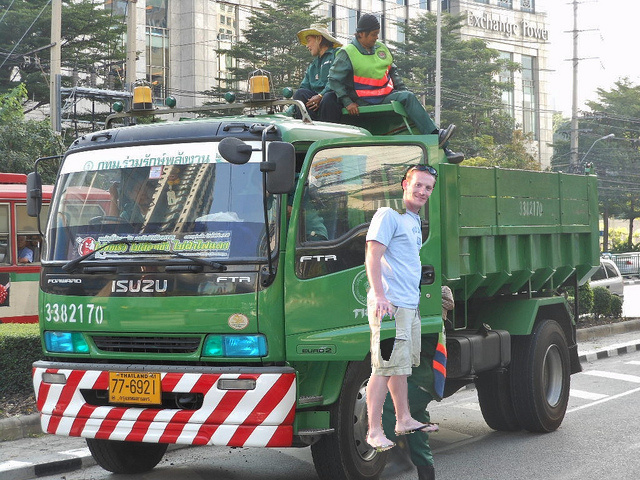}
\end{minipage}
\begin{minipage}{0.160\textwidth}
\includegraphics[width=\textwidth]{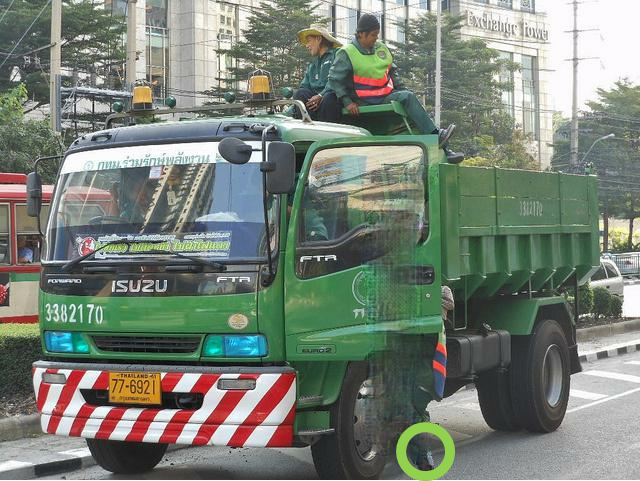}
\end{minipage}
\begin{minipage}{0.160\textwidth}
\includegraphics[width=\textwidth]{./fig/018837/GT.png}
\end{minipage}
\begin{minipage}{0.160\textwidth}
\includegraphics[width=\textwidth]{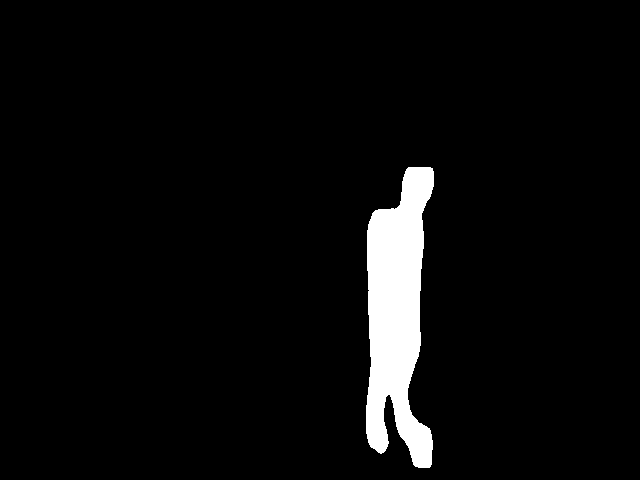}
\end{minipage}
\begin{minipage}{0.160\textwidth}
\includegraphics[width=\textwidth]{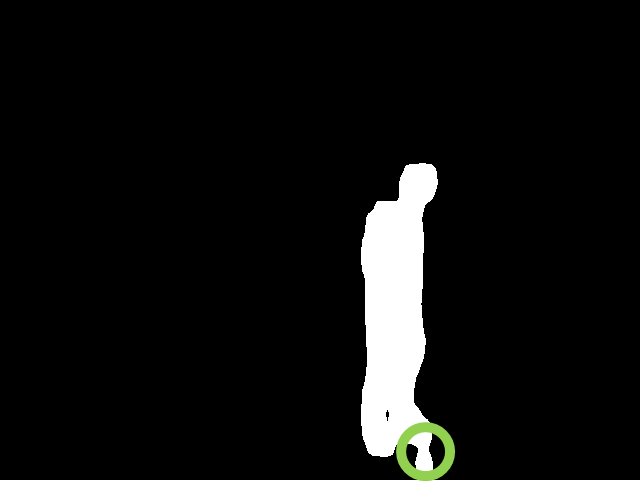}
\end{minipage}\\
\vspace*{3mm}
\begin{minipage}{0.160\textwidth}
\includegraphics[width=\textwidth]{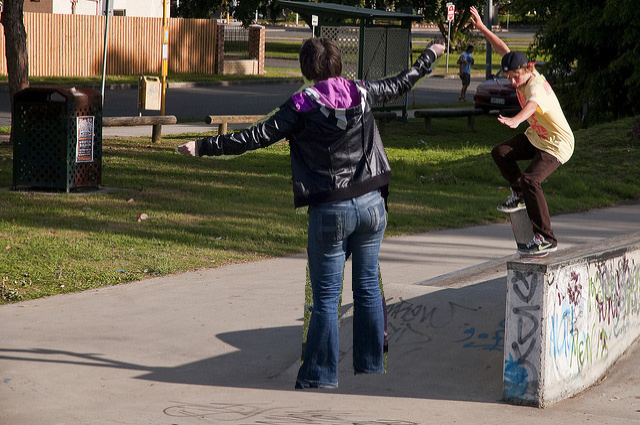}
\centering{(a) Input}
\end{minipage}
\begin{minipage}{0.160\textwidth}
\includegraphics[width=\textwidth]{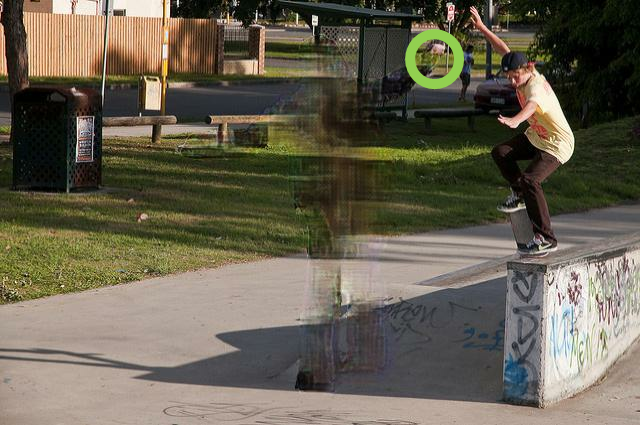}
\centering{(b) Ours}
\end{minipage}
\begin{minipage}{0.160\textwidth}
\includegraphics[width=\textwidth]{./fig/013546/GT.png}
\centering{(c) GT}
\end{minipage}
\begin{minipage}{0.160\textwidth}
\includegraphics[width=\textwidth]{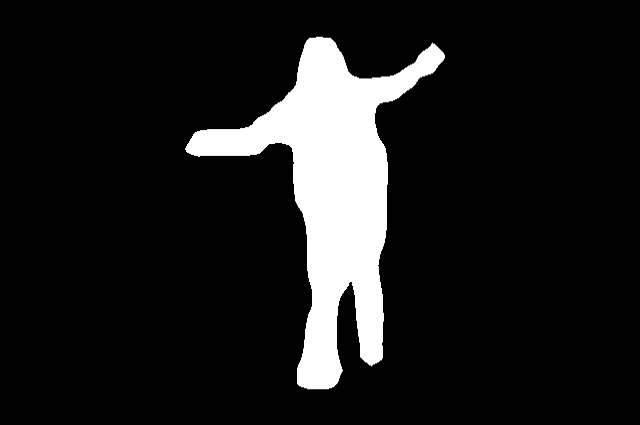}
\centering{(d) Mask}
\end{minipage}
\begin{minipage}{0.160\textwidth}
\includegraphics[width=\textwidth]{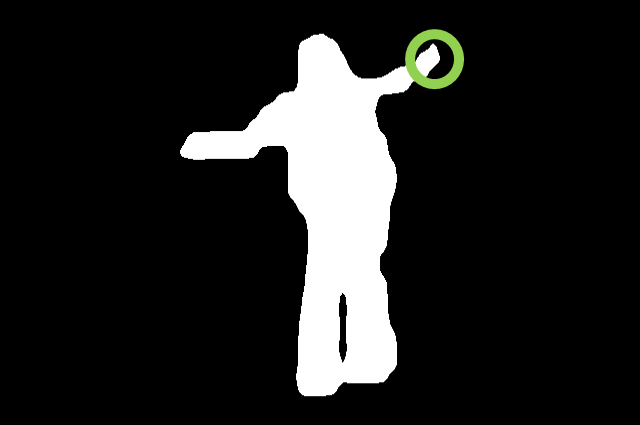}
\centering{(e) Mask by ours}
\end{minipage}\\
\caption{Visual results (failure case). The input and ground-truth images are shown in (a) and (c), respectively. The mask and inpainted images of our method with MEL are in (e) and (b), respectively. The mask generated by PanopticFPN
is shown in (d) for comparison.}
 \label{fig:results_expt4_fail}
\end{center}
\end{figure*}

\subsection{Dataset and Implementation Details}
\label{subsection:exp_details}

We used 118,287 and 5,000 images in the COCO dataset~\cite{LinMBHPRDZ14} for training and test, respectively.
From these images, the paired imageset is generated, as described in Sec.~\ref{subsection:method_dataset}.
As objects selected for generating the paired images, all segments of the person class are used in all experiments shown in this paper.
Panoptic segmentation labels (i.e., pixelwise class annotations), which are required for the paired image generation, are also included in this dataset.
With these annotations, object regions are extracted with COCO API~\cite{lin2016coco}.

$\lambda_{EG}=1$ and $\lambda_{EF}=10$ in Eq.~(\ref{eq:loss_ec_e}), $\lambda_{SG}=0.1$ and $\lambda_{SC}=1$, in Eq.~(\ref{eq:loss_seg}), and $\lambda_{IG}=\lambda_{IF}=0.1$, $\lambda_{IS}=250$, and $\lambda_{IR}=1$ in Eq.~(\ref{eq:loss_ec_i}).
These values in Eq.~(\ref{eq:loss_ec_e}) and Eq.~(\ref{eq:loss_ec_i}) come from~\cite{NazeriNJQE19}.
Those in Eq.~(\ref{eq:loss_seg}) are determined empirically in our experiments.

The implementation of PanopticFPN~\cite{DBLP:conf/cvpr/KirillovHGRD19} is provided by the Detectron2~\cite{wu2019detectron2} library.
The pretrained model is also provided by the Detectron2~\cite{wu2019detectron2}.
In accordance with PanopticFPN, Adam~\cite{KingmaB14} is used as an optimizer.
Its learning rate and mini-batch size are $10^{-5}$ and 4, respectively.

Experimental results are quantitatively evaluated with PSNR and SSIM, in which a higher value is better.
In Tables~\ref{table:expt_3},
\ref{table:exp_1},
\ref{table:exp_2}, 
and \ref{table:exp_7},
the results are shown separately based on the ratio of the pixels of each image to the mask region.

The mean inference time is 0.62 secs with A100.

\subsection{Random masks vs. Object Masks}
\label{subsection:exp_1}

The effect of bridging domains between training and test images is validated.
Our inpainting network is trained independently with either of the two mask types, namely the object mask or the random-pattern mask~\cite{LiuRSWTC18}.
The object mask is provided by image segmentation~\cite{DBLP:conf/cvpr/KirillovHGRD19}, which is pretrained independently of the inpainting network.
The inpainting networks trained with these two mask types are compared as shown in Table~\ref{table:exp_1}.
As shown in Table~\ref{table:exp_1}, The network trained with the object masks is better than the one trained with the random masks in all cases.
In particular, the performance gap is bigger in larger mask sizes.
For example, the PSNR gaps are 0.24 and 1.40 in the mask ratios of 0-10\% and 30-40\%, respectively.

\subsection{The Number of Paired Images for Finetuning}
\label{subsection:exp_2}

Most inpainting networks are trained with random masks.
Therefore, finetuning from those pretrained networks efficiently trains the network with fewer images with object masks.
In Table~\ref{table:exp_2}, we can see that the network pretrained with random masks is improved by finetuning in all cases.
However, the performance is almost saturated between 20k and 100k images.
In particular, in terms of SSIM, the performance is not almost changed by finetuning because the performance gap between the networks trained with random and object masks is not significant in SSIM, as shown in Table~\ref{table:exp_1}.

\subsection{Weight in the Mask Expansion Loss}
\label{subsection:exp_4}

For our experiments, the best $\alpha$ was determined empirically.
As shown in Table~\ref{table:expt5_2}, $\alpha = 0.03$ is the best value.
$\alpha = 0.03$ was used in experiments shown in Sec.~\ref{subsection:exp_7}.

\subsection{Comparison with SOTA Inpainting Methods}
\label{subsection:exp_7}

In Table~\ref{table:exp_7}, our proposed method is compared with SOTA inpainting methods:
G\&L~\cite{DBLP:journals/tog/IizukaS017},
Contextual~\cite{DBLP:conf/cvpr/Yu0YSLH18},
Gated~\cite{DBLP:conf/iccv/YuLYSLH19},
EdgeC~\cite{NazeriNJQE19},
Hi-fill~\cite{DBLP:conf/cvpr/YiTAJX20},
Bat-fill~\cite{DBLP:conf/mm/YuZWPCLMXM21},
Wavefill~\cite{DBLP:conf/iccv/YuZLPMXM21},
RFR~\cite{DBLP:conf/cvpr/LiWZDT20}, 
Boundary~\cite{yamashita2021},
ZITS~\cite{DBLP:conf/cvpr/DongCF22},
MAT~\cite{DBLP:conf/cvpr/LiLZQWJ22}, and
LDM~\cite{DBLP:conf/cvpr/RombachBLEO22}.
The official codes of all of these SOTA methods are publicly available.
The mask image provided by PanopticFPN is fed into each of these SOTA methods.
For ablation study, our method without the Mask Expansion Loss (MEL) is also evaluated.
In addition to PSNR and SSIM as image reconstruction measures, LPIPS~\cite{DBLP:conf/cvpr/ZhangIESW18} and FID~\cite{DBLP:conf/nips/HeuselRUNH17} as perceptual image metrics are used for detailed evaluation.

While Hi-fill gets better results in some metrics, our method is superior to the SOTA methods in many metrics.
In terms of PSNR and SSIM, our method w/ MEL overall outperforms the SOTA methods.
For example, the PSNR differences between our method w/ MEL and the best of all the SOTA methods are 0.64 ($= 31.07-30.43$), 4.11 ($= 24.71 - 20.60$), 7.51 ($= 25.62 - 18.11$), and 9.87 ($= 26.50 - 16.63$) in the mask ratios 0--10\%, 10--20\%, 20--30\%, and 30--40\%, respectively.
In comparison between ours w/ and w/o MEL, MEL overall improves PSNR, SSIM, and LPIPS but degrades FID.
While LPIPS and FID in smaller masks are better in Hi-fill than our methods (i.e., LPIPS in the mask ratios 0--10\% and 10--20\% and FID in 0--10\%), the gaps between Hi-fill and ours are not large.

Figure~\ref{fig:results_expt4_success} shows visual examples.
With the mask image (p) estimated by PanopticFPN, each input image (a) is fed into the inpainting network.
As shown in Fig.~\ref{fig:results_expt4_success} (b)--(l), the SOTA methods fail to remove the target region of object removal.
While Hi-fill gets the best results in both examples, small edge parts still remain, as enclosed by the red circles in Fig.~\ref{fig:results_expt4_success} (f).
These unsuccessful results are obtained because of the smaller masks (p) given by PanopticFPN trained independently of the inpainting network.
On the other hand, our methods (m) and (n) can remove most parts of the target object by employing the masks (q) and (r).
By comparing (m), (n), and GT image (o) more in detail, the whitish object color remains in the result of our method without MEL (m) in the upper example as enclosed by the yellow circle, and blurs and bleeding in (m) are worse than (n) in the lower example as enclosed by the blue ellipse.

Our MEL sometimes fails to correctly expand the mask, as shown in Fig.~\ref{fig:results_expt4_fail}.
In the upper example, the mask generated by our method with MEL is over-expanded, as enclosed by the green circle in (e).
This over-expansion produces the blurred region in (b), as enclosed by the green circle.
In the lower example, the mask is not sufficiently expanded to cover the right hand of the person of a removal target, as enclosed by the green circle in (e).
These results reveal that it is not easy to further optimize mask expansion, which is one of the important future tasks.


\section{Concluding Remarks}
\label{section:conclusion}

This paper proposed a joint-learning framework consisting of the segmentation and inpainting networks for object removal.
The two networks are jointly trained so that the object masks extracted by the segmentation network are optimized for the inpainting task.
In addition, our mask expansion loss minimally expands the mask to cover the object region.

Future work includes the following aspects.
While only person regions are inpainted in this paper, a variety of objects can be targeted by our method.
One more issue is how to copy and paste objects for synthesizing paired images.
While random copy-and-paste is effective for data augmentation~\cite{DwibediMH17,DBLP:conf/cvpr/GhiasiCSQLCLZ21}, object pasting based on the context~\cite{DBLP:conf/eccv/DvornikMS18} is validated.
Such context-based object pasting is one of the future directions for improving the proposed method.

This work was supported by JSPS KAKENHI Grant Number 22H03618.


\bibliographystyle{unsrt}
\bibliography{main}

\end{document}